\documentclass[letterpaper]{article}
\usepackage[preprint]{aaai2027}
\usepackage[hyphens]{url}
\usepackage{graphicx}
\usepackage{natbib}
\usepackage{caption}
\usepackage{algorithm}
\usepackage{algorithmic}
\usepackage{booktabs}
\usepackage{amsmath}
\usepackage{amssymb}
\usepackage{array}

\newcommand{\method}{TEPA}
\newcommand{\Active}{\mathcal{A}}
\newcommand{\Memory}{\mathcal{M}}
\newcommand{\Archive}{\mathcal{R}}
\newcommand{\Retrieve}{\rho}
\newcommand{\Phases}{\Phi}
\newcommand{\conflict}{\operatorname{conflict}}
\newcommand{\succrate}{S}
\newcommand{\mpi}{\operatorname{MPI}}

\title{TEPA: Revoking Stale Memories for Conflict-Robust Language Agents}

\author{
Yan Zhou\textsuperscript{\rm 1},
Yue Ouyang\textsuperscript{\rm 1},
Kaiyang Zheng\textsuperscript{\rm 1},
Suncheng Xiang\textsuperscript{\rm 2}\corresponding
}
\affiliations{
\textsuperscript{\rm 1}School of Mathematics and Statistics, Changsha University of Science and Technology, Changsha, China\\
\textsuperscript{\rm 2}School of Biomedical Engineering, Shanghai Jiao Tong University, Shanghai, China. E-mail: \texttt{xiangsuncheng17@sjtu.edu.cn}
}

\newcommand{\Risk}{\mathcal{L}}
\begin{document}

\maketitle

\vspace{-1em}
\begin{center}
{\footnotesize\textcolor{gray}{This paper is a preprint only; the final published version, if any, shall prevail.}}
\end{center}
\vspace{0.5em}

\begin{abstract}
Long-term memory enables language agents to reuse past facts, preferences, and task experience. Persistence also creates a central falsifiability problem: when the world changes, stale memories can remain retrievable and pollute the prompt. We characterize this failure mode as \emph{memory pollution}: degradation caused by active memories that newer conflicting evidence has superseded. We introduce \method{}, a revocable evidence-memory mechanism that makes validity an explicit state of memory. \method{} represents observations as keyed precedents and revokes active precedents when fresh evidence contradicts them under the same key, allowing retrieval to draw from current evidence while preserving revoked history for audit. Across controlled hidden-regime drift, real file-backed executable drift, and preference-update streams, revocation prevents stale active memory from remaining in the retrieval set after reversal. In controlled drift over 50 seeds, append-only and last-write-wins memory fell below no memory during full reversal (append-only and last-write-wins both 0.210, no memory 0.309, \method{} 0.950), and the same pattern reproduced under real file execution (append-only 0.203, no memory 0.298, \method{} 0.950). On clean MemoryAgentBench SH-6k, \method{} matches a strong last-write-wins cache, confirming that current-key replacement is the decisive operation for single-hop fact consolidation. Boundary tests on multi-hop and very long-context MemoryAgentBench settings expose retrieval-chain and context-selection bottlenecks beyond fact-level validity tracking. Together, these results establish lifecycle revocation as a core memory operation for agents that must falsify, audit, and later re-promote evolving knowledge.
\end{abstract}

\section{Introduction}

Language agents increasingly rely on long-term memory to reuse user facts, reflections, task experience, and skills across interactions \citep{shinn2023reflexion,zhong2024memorybank,park2023generative,zhao2024expel,wang2023voyager,xu2026mem,chhikara2025mem0}. This persistence is useful, yet it creates a falsifiability problem. Once retrieved, a memory becomes prompt evidence, and stale evidence can dominate the current interaction.

\begin{figure}[!t]
\centering
\includegraphics[width=0.94\columnwidth]{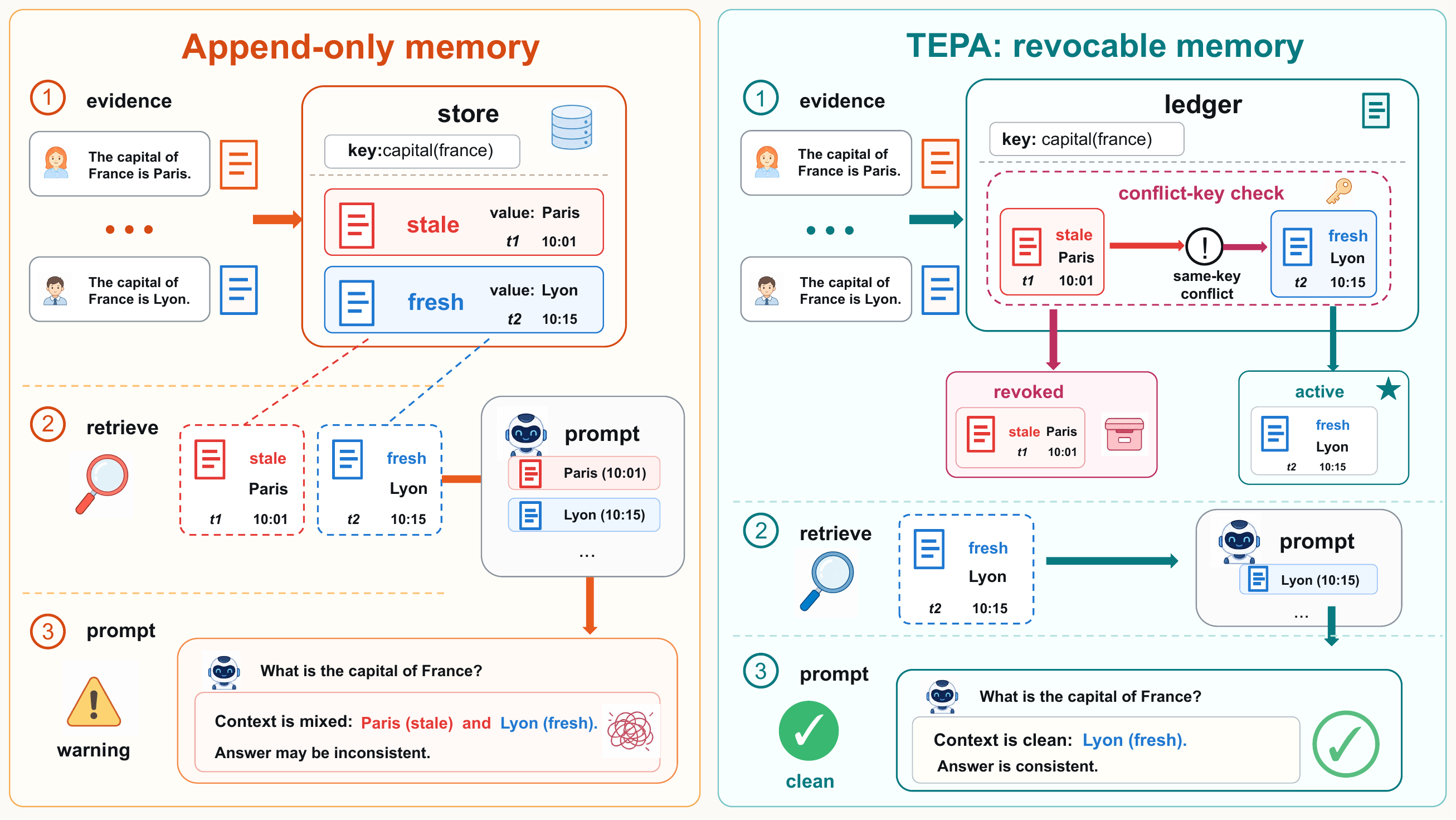}
\caption{Conceptual overview of revocable evidence memory. Append-only memory keeps stale and fresh evidence active under the same key, allowing both to be retrieved into the agent prompt. \method{} performs a conflict-key check, moves superseded precedents into a revoked archive, and retrieves active current evidence.}
\label{fig:concept}
\end{figure}

Agent memory systems have become increasingly effective at growing reusable experience. Reflexion appends lessons, MemoryBank accumulates conversational memories, Generative Agents rank memory streams, ExpeL stores experiential rules, and recent systems add richer memory construction and retrieval \citep{shinn2023reflexion,zhong2024memorybank,park2023generative,zhao2024expel,pan2025memory,xu2026mem,kang2025memory,yu2026agentic}. Falsification remains less explicit: when later evidence contradicts an old memory, the obsolete item may remain retrievable alongside the current fact.

We characterize the resulting degradation as \emph{memory pollution}. It is related to concept drift \citep{gama2004learning,bifet2007learning,gama2014survey}, but has a distinct operational form in language-agent memory: the stale object is textual evidence inserted into the prompt. We focus on \emph{stale-conflict consolidation}, where later observations supersede earlier values under the same key and evaluation asks for the current answer. This setting captures changed preferences, updated entity attributes, and hidden tool-regime shifts, and aligns with recent long-term memory benchmarks \citep{maharana2024evaluating,wu2024longmemeval,hu2025evaluating,wu2026longmemeval}. It lets us pose a direct systems question: can persistent memory remain useful when its own evidence must be falsified?

We introduce \method{}, a revocable evidence-memory mechanism. \method{} converts observations into keyed precedents, assigns each precedent an explicit lifecycle state, and retrieves active precedents. When fresh evidence conflicts with an active precedent under the same key, \method{} performs a local state transition that removes the stale precedent from ordinary retrieval while keeping it available for audit. A trial-validated variant, \method{}-Full, tests candidate precedents before promotion when executable validation is available. Figure~\ref{fig:concept} summarizes the mechanism.

Our evaluation first tests whether memory can become worse than no memory when stale evidence remains active, then measures whether revocation prevents that failure. Across the controlled, executable, and preference-update stale-conflict streams, append-only and key-cache baselines degrade under reversal while revocable lifecycle memory remains robust. MemoryAgentBench SH-6k situates this mechanism within standard fact-consolidation benchmarks, and the multi-hop and very long-context variants identify the next bottlenecks for retrieval-chain construction.

Our contributions are:
\begin{itemize}
    \item We formulate memory pollution as a validity-state failure in long-term language-agent memory, and define a phase-wise pollution index relative to no memory.
    \item We propose \method{}, a conflict-keyed evidence-memory mechanism with explicit precedent states and revocation.
    \item We evaluate \method{} across four conflict settings---controlled, executable, preference-update, and MemoryAgentBench---against mechanism baselines and ablations with paired statistical tests.
\end{itemize}

\section{Related Work}

\paragraph{Memory systems for agents.}
Language-agent memory systems store different forms of experience for later reuse. Reflexion stores verbal lessons from failed attempts \citep{shinn2023reflexion}; MemoryBank maintains long-term conversational memories with forgetting-inspired updates \citep{zhong2024memorybank}; Generative Agents retrieve memory-stream records by recency, importance, and relevance \citep{park2023generative}; ExpeL distills experience into reusable rules \citep{zhao2024expel}. Recent systems such as SeCom, A-MEM, Mem0, MemoryOS, and broader agentic-memory frameworks improve memory construction, retrieval, evolution, and deployment \citep{pan2025memory,xu2026mem,chhikara2025mem0,kang2025memory,yu2026agentic}. These works establish the value of persistent memory. Our work studies the complementary operation that these systems leave comparatively implicit: deactivating a previously useful memory after newer conflicting evidence supersedes it.

\paragraph{Memory benchmarks and stale conflict.}
Long-term memory benchmarks increasingly test whether agents can use information accumulated over long histories. LoCoMo evaluates long-term conversational memory \citep{maharana2024evaluating}; LongMemEval and LongMemEval v2 study assistant memory over extended interactions \citep{wu2024longmemeval,wu2026longmemeval}; MemoryAgentBench evaluates memory through incremental interactions and includes conflict-resolution settings \citep{hu2025evaluating}. Recent stale-memory and agent-memory benchmarks further emphasize validity detection, provenance, interdependent multi-session tasks, and broader memory competencies \citep{chao2026stale,joshi2026eywa,he2026memoryarena,tan2025membench}. Motivated by these benchmarks, we study explicit lifecycle updates as the memory operation underlying single-hop stale-conflict consolidation.

\paragraph{Relation to retrieval and drift.}
Retrieval-augmented generation and long-context methods improve access to external evidence \citep{lewis2020retrieval,karpukhin2020dense,liu2024lost}, while concept-drift and continual-learning methods study adaptation under changing distributions \citep{bifet2007learning,gama2014survey,kirkpatrick2017overcoming,lopez2017gradient}. \method{} operates at the memory-state layer, deciding which explicit memories remain active before retrieval inserts them into the prompt. Additional related-work discussion is provided in the supplementary material.

\section{Problem Formulation}

\paragraph{Episode stream.}
Let an agent interact with an episode stream $E=(e_1,\ldots,e_T)$. Each episode is
\[
e_t=(c_t,x_t,z_t),
\]
where $c_t$ is the visible decision context, $x_t$ is the post-action evidence item used to update memory, and $z_t$ is a hidden benchmark regime. Before acting, method $m$ observes $c_t$, retrieves from memory, and selects $a_t^{(m)}$. It then observes $y_t^{(m)}\in\{0,1\}$ together with $x_t$ and updates memory. The hidden regime remains evaluator-side. A phase function $\phi(t)\in\Phases$ maps each episode to a phase, where $\Phases=\{\text{stable},\text{light},\text{reversal},\text{return}\}$ in our drift benchmarks.

\paragraph{Memory and retrieval.}
The agent maintains memory $\Memory_t$ after episode $t$. At decision time it retrieves a subset $\Retrieve(c_t,\Active_t)\subseteq \Active_t$, where $\Active_t\subseteq\Memory_t$ is the active memory set. A method policy $\pi_m$ then selects $a_t^{(m)}$ conditioned on $(c_t,\Retrieve(c_t,\Active_t))$. Append-only memory keeps every stored item active. Revocable memory separates active precedents $\Active_t$ from revoked precedents $\Archive_t$.

\paragraph{Conflict keys.}
Each evidence item is mapped to a key-value pair $(k_t,v_t)$ by key extractor $\kappa$ and value extractor $\nu$: $k_t=\kappa(x_t)$ and $v_t=\nu(x_t)$. The key identifies what the evidence is about, such as a user preference slot or a tool-regime slot; the value is the current claim under that key. Two evidence items conflict when they share a key but assert incompatible values:
\[
\conflict(x_i,x_j)=\mathbb{1}[\kappa(x_i)=\kappa(x_j)\wedge \nu(x_i)\neq\nu(x_j)].
\]
Stale-conflict consolidation asks the agent to answer using the current value for each key.
The extractors $\kappa$ and $\nu$ are deterministic in the reported benchmarks; the supplementary material audits sensitivity when keys are noisy.

\paragraph{Success rate and pollution.}
For method $m$ and phase $\phi$, let
\[
\succrate_{m,\phi}=\frac{1}{|\mathcal{T}_\phi|}\sum_{t\in \mathcal{T}_\phi} y_t^{(m)}
\]
where $\mathcal{T}_\phi=\{t:\phi(t)=\phi\}$ and $y_t^{(m)}$ is the success observed after method $m$ acts. We define the phase-wise memory pollution index of method $m$ relative to no memory as
\[
\mpi_{m,\phi}=\frac{\succrate_{\mathrm{NoMem},\phi}-\succrate_{m,\phi}}
{\succrate_{\mathrm{NoMem},\phi}}.
\]
Positive $\mpi$ means the method is worse than no memory in that phase. We use it as a reversal-phase diagnostic alongside absolute success. A key failure case is $\succrate_{m,\phi}<\succrate_{\mathrm{NoMem},\phi}$ for an append-only memory method during reversal.

\section{Method}

Table~\ref{tab:notation} summarizes the notation used in the formulation and method.

\begin{table*}[!t]
\centering
\footnotesize
\setlength{\tabcolsep}{3.5pt}
\caption{Notation used in the formulation and method.}
\label{tab:notation}
\renewcommand{\arraystretch}{0.92}
\begin{tabular}{@{}p{0.12\textwidth}p{0.18\textwidth}p{0.62\textwidth}@{}}
\toprule
\textbf{Symbol} & \textbf{Name} & \textbf{Definition} \\
\midrule
\multicolumn{3}{@{}l}{\textit{Core objects}} \\
\midrule
$e_t$ & Episode & Interaction at time $t$, consisting of visible context, post-action evidence, and hidden regime. \\
$c_t$ & Context & Visible task or query context supplied to the agent. \\
$x_t$ & Evidence & Fact, preference, tool outcome, or task experience observed after acting and used to update memory. \\
$a_t^{(m)}$ & Action & Agent answer or tool/action choice produced by method $m$. \\
$y_t^{(m)}$ & Outcome & Binary task success indicator observed after method $m$ acts. \\
$z_t$ & Hidden regime & Benchmark state controlling the current correct value; unobserved by the agent. \\
$\phi(t)$ & Phase & Drift phase containing episode $t$. \\
$\Memory_t$ & Memory & All stored precedents after episode $t$. \\
$\Active_t$ & Active set & Precedents eligible for retrieval. \\
$\Archive_t$ & Revoked archive & Revoked precedents excluded from normal retrieval. \\
$\Retrieve$ & Retriever & Function returning active precedents for a context. \\
$\pi_m$ & Policy & Method-specific decision rule mapping context and retrieved precedents to an action or answer. \\
\midrule
\multicolumn{3}{@{}l}{\textit{Conflict mechanics}} \\
\midrule
$\kappa,\nu$ & Extractors & Functions mapping evidence to conflict key $\kappa(x)$ and asserted value $\nu(x)$. \\
$p$ & Precedent & Keyed memory record $(k,v,s,f,\sigma,\tau)$. \\
$k,v$ & Key and value & Conflict key and asserted value. \\
$s,f$ & Evidence counts & Support and failure/conflict counts for a precedent. \\
$\sigma$ & State & Lifecycle state in \{\textsc{Hypothesis}, \textsc{Active}, \textsc{Revoked}\}. \\
$q(p)$ & Posterior mean & Beta-Bernoulli estimate of current validity for precedent $p$. \\
$\alpha,\beta$ & Prior counts & Beta prior parameters used in $q(p)$; set to one in all experiments. \\
$H$ & Held-out tasks & Trial-validation task set used by \method{}-Full. \\
\midrule
\multicolumn{3}{@{}l}{\textit{Validation and metrics}} \\
\midrule
$\theta_{\rm prop},\theta_{\rm rev},$\\$\theta_{\rm pro},\theta_{\rm rec}$ & Thresholds & Proposal, revocation, promotion, and recent-success thresholds. \\
$n_{\min},$\\$n_{\rm rec}$ & Observation counts & Minimum total and recent-window observation counts for revocation checks. \\
$\succrate_{m,\phi}$ & Success rate & Mean task success of method $m$ in phase $\phi$. \\
$\mpi_{m,\phi}$ & Pollution index & Relative gap between no memory and method $m$ in phase $\phi$. \\
\bottomrule
\end{tabular}
\end{table*}

\subsection{Precedents with Explicit Validity State}

Algorithm~\ref{alg:tepa} gives the full memory-update procedure.

\method{} represents memory as a set of precedents. A precedent is
\[
p=(k,v,s,f,\sigma,\tau),
\]
where $k$ is the conflict key, $v$ is the asserted value or action pattern, $s$ and $f$ are support and conflict counts, $\sigma$ is the lifecycle state, and $\tau$ is the creation time. Retrieval is conditioned on the active state, $\sigma=\textsc{Active}$.

The retriever used in drift benchmarks returns active precedents whose key matches the current task key, ordered by recency within that key. In MemoryAgentBench, all methods use the same lexical top-$k$ answer context after their memory-state update. The posterior mean of a precedent is estimated with a Beta-Bernoulli model:
\[
q(p)=\frac{s_p+\alpha}{s_p+f_p+\alpha+\beta},
\]
with $\alpha=\beta=1$ in all experiments. We use this posterior as a transparent state variable for lifecycle transitions; the update rule asks whether recent evidence still supports an active precedent.

\subsection{Lifecycle Update}

When an episode produces evidence $(k_t,v_t)$, \method{} compares it with active precedents sharing key $k_t$. If an active precedent $p$ asserts the same value, its support count $s_p$ increases. If it asserts an incompatible value, its conflict count $f_p$ increases. A precedent leaves the active set when either its posterior mean falls below $\theta_{\rm rev}$ after $n_{\min}=5$ total observations or, once at least $n_{\rm rec}=3$ same-key outcomes are available, its recent success rate falls below $\theta_{\rm rec}=0.34$. Fresh evidence can later promote a new precedent under the same key.

The lifecycle states serve different purposes. \textsc{Hypothesis} stores candidate precedents before promotion. \textsc{Active} precedents are retrievable. \textsc{Revoked} marks a precedent as contradicted by fresher same-key evidence. Revoked memory remains in the archive for audit while leaving the active retrieval set.

\subsection{Trial-Validated Promotion}

\method{}-Full adds trial-by-execution before promotion. Given a candidate precedent $p$ and a held-out task set $H$, a task executor evaluates whether injecting $p$ improves reward on support tasks, avoids harm on counterfactual tasks, and keeps unrelated domains uncontaminated. With the default trial budget of 3, the executor allocates one support, one counterfactual, and one contamination check. A candidate is promoted after at least 60\% of trials are positive and at least one support trial succeeds. The trial variant matches default \method{} on controlled drift and improves robustness on preference updates, where validation can catch incorrect candidates before activation.

\begin{algorithm}[t]
\caption{\method{} memory update}
\label{alg:tepa}
\begin{algorithmic}[1]
\REQUIRE Episode stream $E$, active memory $\Active_0=\emptyset$, archive $\Archive_0=\emptyset$
\FOR{$t=1,\ldots,T$}
    \STATE Retrieve $R_t \leftarrow \Retrieve(c_t,\Active_{t-1})$
    \STATE Execute policy $a_t^{(m)} \sim \pi_m(c_t,R_t)$ and observe $y_t^{(m)},x_t$
    \STATE Extract $k_t=\kappa(x_t)$ and $v_t=\nu(x_t)$
    \FOR{each $p\in\Active_{t-1}$ with $p.k=k_t$}
        \IF{$p.v=v_t$}
            \STATE $p.s \leftarrow p.s+1$
        \ELSE
            \STATE $p.f \leftarrow p.f+1$
            \IF{$q(p)<\theta_{\rm rev}$ after $n_{\min}$ observations or (recent count $\ge n_{\rm rec}$ and recent success $<\theta_{\rm rec}$)}
                \STATE move $p$ from $\Active$ to $\Archive$ with state \textsc{Revoked}
            \ENDIF
        \ENDIF
    \ENDFOR
    \IF{accumulated supports for $(k_t,v_t)$ $\geq \theta_{\rm prop}$}
        \STATE create or update candidate $p'$ in state \textsc{Hypothesis}
        \IF{$q(p')\geq\theta_{\rm pro}$ or $p'$ passes trial validation}
            \STATE set $p'.\sigma=\textsc{Active}$ and add $p'$ to $\Active$
        \ENDIF
    \ENDIF
\ENDFOR
\end{algorithmic}
\end{algorithm}

\subsection{Design Contrast}

Append-only memory keeps all stored evidence active. Temporal recency and sliding windows reduce exposure to old evidence, sometimes discarding useful memories or retaining stale recent ones. Reactive forgetting clears memory after detected failure at a global level. Oracle reset knows phase boundaries and serves as a diagnostic upper bound. \method{} makes the unit of update a keyed precedent; each revocation affects active precedents sharing the conflicting key while preserving other keys for retrieval.

\paragraph{Naming convention.}
Unless qualified, \method{} denotes the default revocable variant (\method{}-Rev in Table~\ref{tab:mab_sh6k}). \method{}-NoRev is an ablation that omits revocation. \method{}-Full is the trial-validated variant used in preference-update experiments.

\paragraph{Theoretical role of revocation.}
A simple risk view closes the link between revocation and the pollution index. Under a single-key supersession model, suppose the current value $v^+$ and a stale value $v^-$ share key $k$, with $v^-\neq v^+$. If append-only retrieval exposes $v^-$ with probability $\xi$ and the downstream policy follows it with probability $\eta$, then append-only memory incurs at least $\xi\eta$ excess 0--1 risk relative to a memory state that revokes $v^-$ while keeping $v^+$ active.

This gives a sufficient condition for memory pollution. Let $b$ denote the benefit, relative to no memory, of any still-current evidence retrieved by append-only memory in a phase. When stale exposure dominates useful reuse, $\xi\eta>b$, append-only memory has higher expected error than no memory, equivalently $\succrate_{{\rm append},\phi}<\succrate_{{\rm NoMem},\phi}$ and $\mpi_{{\rm append},\phi}>0$. Revocation targets exactly the harmful term $\xi\eta$: conditioned on stale exposure and stale following, the answer is wrong by construction; removing only the contradicted same-key value eliminates that exposure while preserving current evidence. The supplementary material gives the full proof and the Beta-statistic consistency note.

\section{Experimental Setup}

\paragraph{Research questions.}
We evaluate four questions. RQ1: Can append-only memory become worse than no memory under hidden reversal? RQ2: Does revocation prevent that failure in controlled and executable environments? RQ3: Do trials improve updates to stale user preferences? RQ4: How does the same lifecycle mechanism transfer to external memory benchmarks?

\paragraph{Benchmarks.}
We use four evaluation settings. Controlled hidden-regime drift and real file-backed executable drift test whether append-only memory becomes harmful under tool-regime reversal. A preference-update stream tests stale user-profile updates. MemoryAgentBench SH-6k provides an external single-hop fact-consolidation benchmark \citep{hu2025evaluating}; MH-6k and long-context variants are reported as boundary settings. Full benchmark details are provided in the supplementary material.

\paragraph{Baselines.}
We use mechanism baselines that isolate alternatives to revocation: no memory, append-only memory, temporal recency, semantic retrieval, sliding window, reactive forgetting, last-write-wins, conflict-aware recency, and oracle reset. These baselines share the same task streams, answer backend, and scoring pipeline, so performance differences reflect memory-state behavior under controlled system conditions. The no-memory baseline measures base task success when no stored evidence is exposed to the prompt. A method falling below this baseline reveals memory pollution: retrieved memory has become an active source of harm. In the drift benchmarks, last-write-wins writes successful same-key experiences, so a stale hidden-regime precedent remains until a new successful value is obtained. In MemoryAgentBench, the stream directly supplies asserted facts, making last-write-wins the strongest clean single-hop replacement baseline. \method{} ablations include no-revocation, threshold variants, recent-window variants, and trial-budget variants. The exact baseline definitions are provided in the supplementary material.

\paragraph{Metrics and statistics.}
For drift and preference benchmarks, the primary metric is phase-wise success rate $\succrate_{m,\phi}$. For MemoryAgentBench, the primary metric is substring exact match. We report bootstrap 95\% confidence intervals across seeds or paired task samples, one-tailed paired permutation tests for directional comparisons, Holm--Bonferroni correction for families of comparisons, and paired effect sizes $d_z$ where applicable. Unless otherwise stated, confidence intervals are 95\%.

\paragraph{Why paired tests.}
The paired design removes a major source of variance: some generated tasks or memory queries are intrinsically easier than others. Comparing methods on matched task indices therefore estimates the effect of the memory mechanism while controlling for task sampling. The reported confidence intervals quantify uncertainty over the paired experimental stream.

\paragraph{Implementation details.}
Controlled, executable, and preference-update benchmarks use deterministic benchmark executors so that the experiments isolate memory behavior from language-model sampling variance. MemoryAgentBench uses a fixed DeepSeek-v4-Flash answer backend under the same prompt and scoring pipeline for all methods. The default \method{} configuration uses proposal threshold $\theta_{\rm prop}=3$, revocation threshold $\theta_{\rm rev}=0.3$, promotion threshold $\theta_{\rm pro}=0.6$, and trial budget 3 for \method{}-Full. All methods are evaluated on the same splits and query orders within each benchmark.

\paragraph{Extraction and leakage controls.}
The reported experiments use schema-based conflict keys for controlled attribution. In controlled and executable drift, $\kappa(x)$ is the tuple of task type, domain, and input type, while $\nu(x)$ is the successful tool pattern recorded by the executor. In preference updates, the key is the preference slot and profile context; in MemoryAgentBench SH-6k, it is parsed from the fact subject and relation template. Deployable methods receive the visible episode stream, while phase labels, future boundaries, and current hidden regimes remain evaluator-side variables. Oracle reset uses privileged phase information and therefore functions as a diagnostic upper bound. These controls make the comparison test memory-state behavior: whether stale same-key evidence remains active after newer conflicting evidence arrives.

\section{Results}

\begin{table}[t]
\centering
\small
\setlength{\tabcolsep}{3.2pt}
\caption{Main evidence summary. Drift and preference rows report full-reversal success; MemoryAgentBench reports substring exact match. Preference uses \method{}-Full; MAB uses \method{}-Rev.}
\label{tab:main_summary}
\begin{tabular}{@{}lcccc@{}}
\toprule
\textbf{Setting} & \textbf{Append} & \textbf{NoMem} & \textbf{LWW} & \textbf{\method{} family} \\
\midrule
Controlled & 0.210 & 0.309 & 0.210 & \textbf{0.950} \\
Executable & 0.203 & 0.298 & 0.203 & \textbf{0.950} \\
Preference & 0.138 & 0.837 & -- & \textbf{0.872} \\
MAB SH-6k & 0.583 & -- & \textbf{0.890} & \textbf{0.890} \\
\bottomrule
\end{tabular}
\end{table}

\subsection{Controlled Drift Exposes Memory Pollution}

\begin{figure*}[t]
\centering
\includegraphics[width=0.68\textwidth]{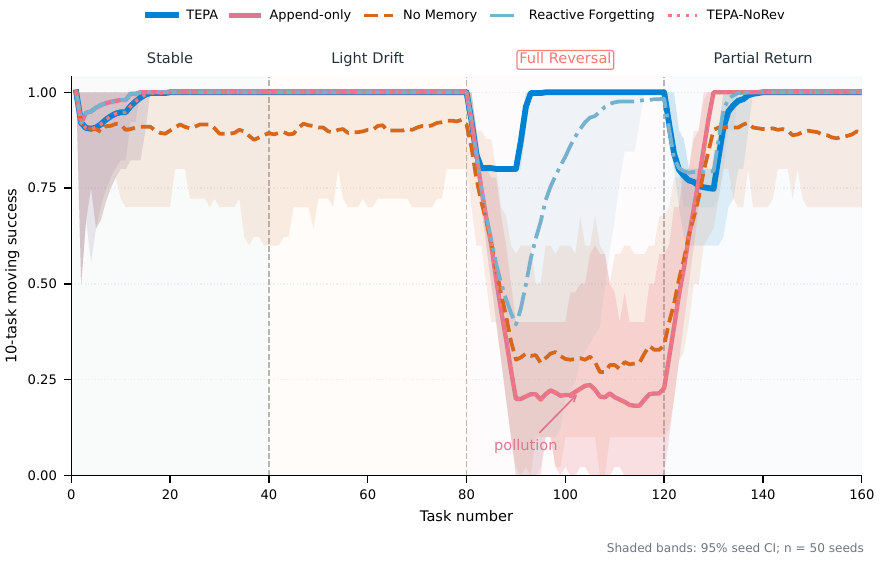}
\caption{Memory pollution under controlled hidden-regime drift. Curves show 10-task moving averages over 50 seeds with 95\% seed confidence bands. During full reversal, append-only memory falls below no memory because the stale precedent remains active; \method{} revokes the stale precedent and stays robust.}
\label{fig:pollution_curve}
\end{figure*}

Table~\ref{tab:main_summary} summarizes the main evidence, and Figure~\ref{fig:pollution_curve} shows the controlled hidden-regime drift result. In stable and light-drift phases, append-only memory performs well because the stored precedent remains correct. During full reversal, the same active memory becomes harmful: append-only, last-write-wins, and conflict-aware recency each fall to 0.210 success, below the no-memory baseline of 0.309. This yields a positive pollution index of 0.318. \method{} maintains 0.950 success in the same phase by revoking stale precedents. The paired \method{}-vs-append-only difference over all tasks is 0.167 (CI [0.162, 0.172], Holm $p<0.001$, $d_z=9.68$); \method{}'s overall advantage over last-write-wins is numerically identical (0.167, CI [0.162, 0.172]), because last-write-wins behaves like append-only in the drift stream under our benchmark design. During full reversal, the \method{}-vs-last-write-wins difference is 0.740 (CI [0.722, 0.756], Holm $p<0.001$).

The identical 0.950 value for \method{} in controlled and executable reversal reflects the shared drift schedule and deterministic memory-update logic. Errors concentrate in the first few same-key trials after reversal, before sufficient contradictory evidence accumulates. Once revocation fires, retrieval contains the current precedent and subsequent same-key tasks are solved deterministically. The 0.950 plateau therefore measures the adaptation latency of the lifecycle rule: the system pays a small transition cost, then recovers a clean active set. This behavior is important because it links the aggregate success rate to a concrete memory event, namely the moment a contradicted precedent leaves retrieval.

The supplementary phase-wise heatmap summarizes the same pattern across methods. The full-reversal column isolates the central phenomenon. Append-only, semantic retrieval, temporal recency, and no-revocation all remain near 0.210 success. Reactive forgetting improves to 0.796, but still trails \method{} at 0.950. Oracle reset reaches 0.812 because it is given phase-boundary resets but uses the same weak base executor. This pattern supports the mechanism claim: stale active memory is the driver of the collapse across methods with different storage policies.

\subsection{The Effect Reproduces with Real File Execution}

\begin{figure}[t]
\centering
\includegraphics[width=0.96\columnwidth]{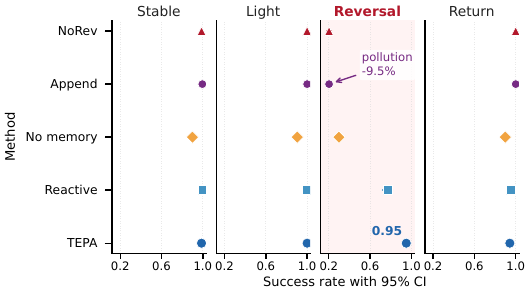}
\caption{Real file-backed tool execution validation. Columns are drift phases and rows are memory methods; points show phase-wise success rates over 50 seeds with 95\% confidence intervals. The highlighted reversal column reproduces memory pollution in executable tasks: append-only memory is below no memory, while \method{} remains robust.}
\label{fig:real_tool}
\end{figure}

The executable benchmark tests whether the controlled result survives contact with real file I/O and tool execution. Figure~\ref{fig:real_tool} reports phase-wise success as a compact dot-interval matrix: columns are drift phases, rows are methods, and horizontal intervals show 95\% confidence intervals over seeds. Under full reversal, append-only and last-write-wins memory reach 0.203 success while no memory reaches 0.298, giving a pollution index of 0.319. \method{} reaches 0.950, outperforming reactive forgetting at 0.771. The \method{}-vs-append-only and \method{}-vs-last-write-wins differences during full reversal are both 0.748 (CI [0.728, 0.768], Holm $p<0.001$). The no-memory-vs-append-only reversal gap is also significant: 0.095 (CI [0.083, 0.107], Holm $p<0.001$, $d_z=2.10$).

\subsection{Preference Updates Require Trial-Validated Revocation}

The preference-update stream tests cases where stale long-term profiles are corrected by fresh session feedback. The supplementary preference-update plot shows that append-only memory collapses under preference reversal: 0.138 full-reversal success versus 0.837 for no memory. Last-write-wins improves over append-only but still falls below no memory overall (0.686 vs.\ 0.837). Augmenting last-write-wins with the same validation rule used by \method{}-Full raises its success to 0.863, yet it still trails \method{}-Full overall by 0.048 (CI [0.026, 0.069], Holm $p=0.0011$), with a larger 0.240 gap concentrated in the full-reversal phase (CI [0.156, 0.330]). \method{}-Full reaches 0.910 overall and is statistically indistinguishable from no memory (difference 0.002, CI [-0.011, 0.014], Holm $p=1.000$). The gap shows that validation and lifecycle state play complementary roles: validation filters candidate promotion, while revocation clears old active preferences.

This setting is deliberately difficult for relevance-based memory because the stale profile remains semantically on-topic after its value has been superseded by newer feedback. The result separates two operations that are easy to conflate. Validation prevents some bad candidates from being promoted; lifecycle revocation removes previously active preferences that remain attractive to retrieval because they match the current query.

\subsection{MemoryAgentBench Single-Hop Conflict Resolution}

MemoryAgentBench SH-6k provides an external stale-conflict consolidation test. Table~\ref{tab:mab_sh6k} reports substring exact match over three query-order seeds and 300 queries. \method{}-Rev reaches 0.890, matching the strong last-write-wins current-key cache and outperforming \method{}-NoRev at 0.630. This result shows that current-key replacement is the decisive operation for clean single-hop fact updates where the latest fact is directly observed. The lifecycle state, archive, and trial-validation machinery become most valuable in the drift and preference settings above, where valid updates must be inferred from interaction outcomes.

\begin{table}[t]
\centering
\small
\setlength{\tabcolsep}{5pt}
\caption{MemoryAgentBench SH-6k single-hop conflict resolution. Metric is substring exact match over three query-order seeds and 300 queries.}
\label{tab:mab_sh6k}
\begin{tabular}{@{}lc@{}}
\toprule
\textbf{Method} & \textbf{Substring EM} \\
\midrule
\multicolumn{2}{@{}l}{\textit{Current-state methods}} \\
\method{}-Rev & \textbf{0.890} \\
Last-write-wins & \textbf{0.890} \\
\method{}-NoRev & 0.630 \\
\midrule
\multicolumn{2}{@{}l}{\textit{Stale-memory baselines}} \\
Append-only & 0.583 \\
Sliding window & 0.580 \\
Reactive forgetting & 0.580 \\
Conflict-aware recency & 0.580 \\
Temporal recency & 0.510 \\
\bottomrule
\end{tabular}
\end{table}

\subsection{Ablations Confirm the Role of Revocation}

\begin{figure}[t]
\centering
\includegraphics[width=0.96\columnwidth]{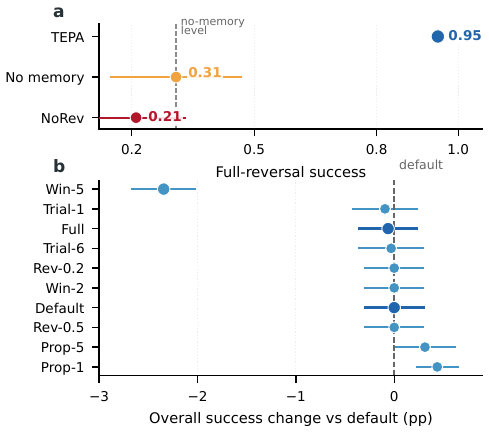}
\caption{Ablation results. Panel (a): full-reversal dot intervals show that removing revocation collapses performance below no memory. Panel (b): hyperparameter variants are shown as percentage-point changes from the default \method{} setting; most thresholds and trial budgets stay close to zero, while a larger recent window slows adaptation.}
\label{fig:ablation}
\end{figure}

Figure~\ref{fig:ablation} separates component necessity from hyperparameter sensitivity. Panel (a) compares full-reversal success directly: removing revocation collapses performance from 0.950 to 0.211, matching append-only behavior and falling below no memory. Panel (b) reports each variant as a percentage-point change from the default \method{} setting. Threshold variants around the default ($\theta_{\rm rev}\in[0.2,0.4]$, $\theta_{\rm pro}\in[0.5,0.7]$) and trial-budget variants (1, 3, 5) remain close to zero, while a larger recent window is the main negative outlier. The ablations identify revocation as the essential component for hidden-regime reversal and show that nearby threshold choices preserve the effect.

\subsection{Boundary Conditions}

Boundary experiments separate memory-state validity from harder retrieval problems. \method{} remains useful on SH-32k (0.680 substring exact match), but drops to 0.040 on MH-6k and 0.000 on SH-262k. Fact-level revocation addresses stale single-hop conflicts; multi-hop relational retrieval and long-context selection emerge as the next architectural challenges. The full boundary table is provided in the supplementary material.

\section{Discussion}

The results point to a design principle for agent memory: persistent memory should track validity alongside relevance. Stale evidence can remain highly relevant to the current query, giving similarity, recency, and larger context a weak signal for supersession. \method{} addresses this at the memory-state layer by removing the contradicted same-key precedent before retrieval while preserving current evidence. The same lifecycle view explains the memory dynamics reported in the supplement: append-only storage grows with every observed fact, windowed or recency-based memory caps exposure by age or position, and \method{} keeps active memory compact by moving contradicted precedents out of the active set while retaining them for audit.

The MemoryAgentBench boundary results sharpen the architectural picture. Revocation repairs the validity state of single-hop facts before they enter the prompt; multi-hop chain construction and very-long-context selection point to retrieval planners as the next systems layer. Two practical consequences follow. Mechanism baselines act as shared-evaluator controls under the same backend and task stream. Archiving revoked precedents supports audit, diagnosis, and later re-promotion while keeping ordinary retrieval focused on currently active evidence.

\section{Limitations}

\method{} assumes useful conflict keys can be extracted from evidence, a natural fit for preference slots, entity attributes, and tool-regime records, and a harder problem for open-ended memories whose conflict relation is implicit. The MH-6k and long-context MemoryAgentBench results mark where chain construction, cross-key composition, and very-long-context selection become dominant beyond fact-level validity tracking. The supplementary key-noise audit quantifies this transition directly: \method{} degrades smoothly as structured key noise increases, while append-only and last-write-wins remain near their reversal failure modes because they lack a lifecycle signal for stale hidden-regime precedents.

\section{Conclusion}

We introduced \method{}, a revocable evidence-memory mechanism for stale-conflict consolidation. Append-only memory can become worse than no memory when stale conflicting evidence remains active, and conflict-keyed revocation prevents this failure in controlled, executable, and preference-update settings. The broader takeaway is operational: persistent memory needs an explicit validity state. A memory item can be relevant, recent, and still wrong once its value has been superseded; revocation removes such evidence from ordinary retrieval while preserving it for validation, audit, and later re-promotion.

This view recasts long-term memory as an inspectable lifecycle governed by validity evidence and state transitions. Each precedent carries a key, evidence, state, and contradiction history, allowing agents to explain why a memory is active, revoked, or ready for re-promotion.

\newpage

\bibliography{tepa_refs_2026_checked}

\section*{Supplementary Material}
\appendix

\section{Appendix A: Mechanism Trace and Memory Dynamics}

\begin{figure*}[h]
\centering
\includegraphics[width=0.78\textwidth]{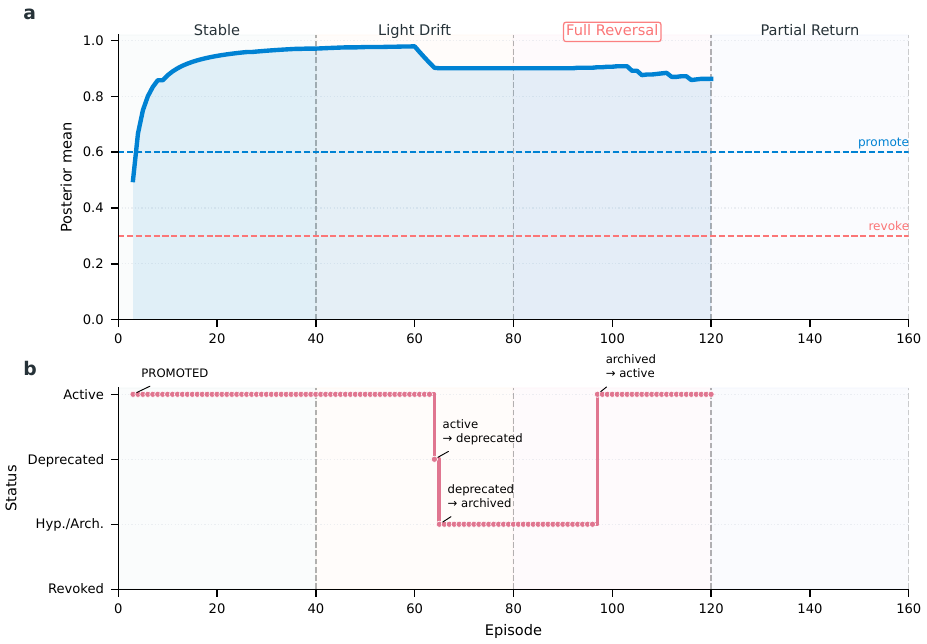}
\caption{Precedent lifecycle trace in controlled drift. The posterior drops under reversal, triggering revocation, and later evidence can support re-promotion.}
\label{fig:lifecycle}
\end{figure*}

\begin{figure}[h]
\centering
\includegraphics[width=\columnwidth]{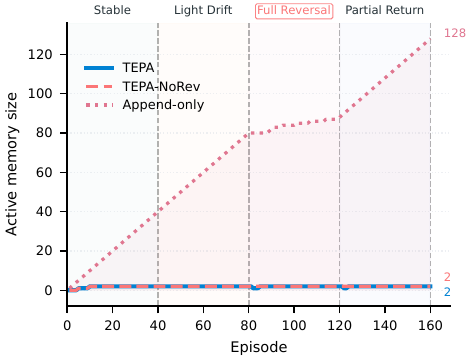}
\caption{Memory-size evolution. Revocation keeps active memory compact while append-only memory accumulates stale evidence.}
\label{fig:memory_size}
\end{figure}

This appendix visualizes the lifecycle mechanism discussed in the main paper. The figures below are audit views of internal memory state: Figure~\ref{fig:lifecycle} follows one representative precedent through promotion, reversal-induced revocation, and later re-promotion, and Figure~\ref{fig:memory_size} shows active memory size over time.

\paragraph{How to read Figure~\ref{fig:lifecycle}.}
The top panel tracks the posterior mean $q(p)$ of a single precedent. The revocation threshold is crossed when repeated counter-evidence arrives during full reversal. The bottom panel records the corresponding discrete lifecycle state, illustrating the archived-state behavior discussed in the main text.

\paragraph{Memory-size interpretation.}
The main paper discusses this interpretation; the figure records the corresponding active-set trace.

\section{Appendix B: Statistical Audit}

\begin{figure}[h]
\centering
\includegraphics[width=\columnwidth]{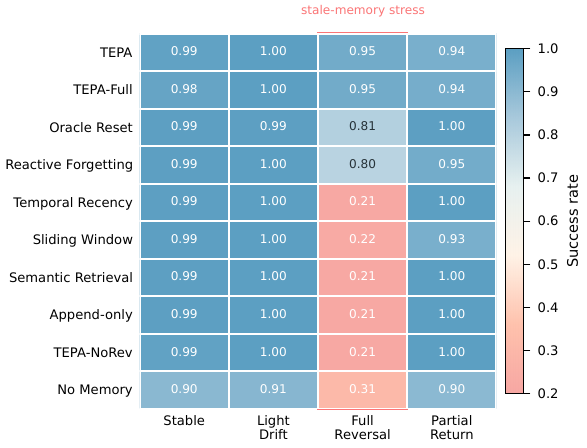}
\caption{Phase-wise success rates in controlled hidden-regime drift. The highlighted reversal phase shows that append-only and no-revocation variants become worse than no memory, whereas revocable memory remains high.}
\label{fig:phase_heatmap}
\end{figure}

Figure~\ref{fig:phase_heatmap} provides the phase-wise method audit referenced in the main text. It makes the same point as the main drift curve with more method-level granularity: the reversal phase is where stale active memory becomes harmful.

\begin{figure}[h]
\centering
\includegraphics[width=\columnwidth]{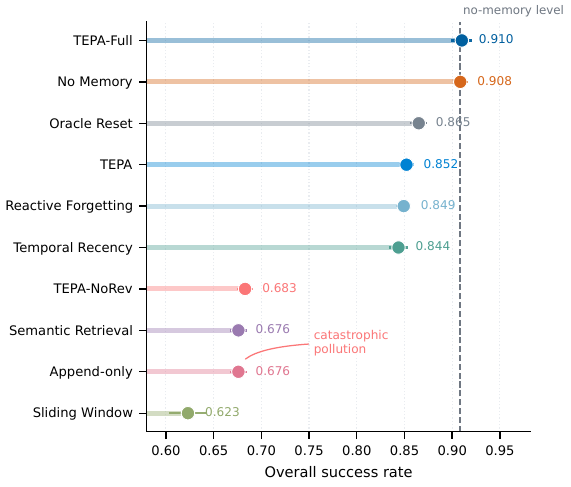}
\caption{Preference-update stream. The horizontal lollipop plot shows overall success over 50 seeds. Append-only memory preserves stale preferences and falls far below no memory. \method{}-Full recovers to the no-memory level by validating updates before promotion.}
\label{fig:preference}
\end{figure}

\begin{figure*}[h]
\centering
\includegraphics[width=0.65\textwidth]{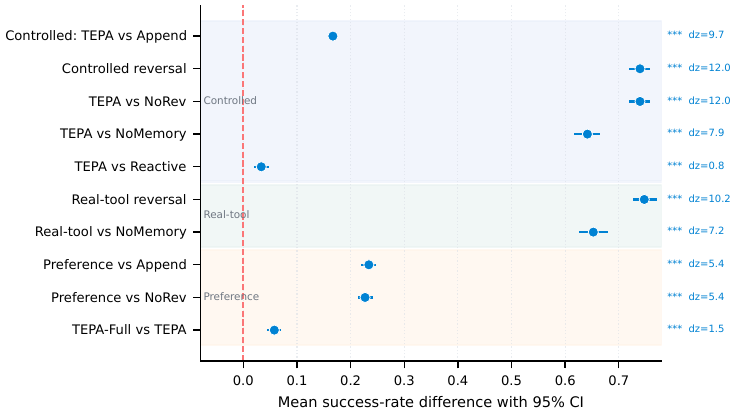}
\caption{Statistical audit summary. Forest plot entries show paired mean differences with 95\% confidence intervals and corrected significance markers.}
\label{fig:stat_audit}
\end{figure*}

The main paper reports only the comparisons needed to interpret each result. This appendix records the statistical protocol used for all reported significance claims. Figure~\ref{fig:stat_audit} visualizes the main paired effects, and Table~\ref{tab:key_stats} lists the numerical comparisons used in the paper. Tests are paired whenever possible: for a fixed seed and task index, each method is evaluated on the same generated task or query. Let $d_i=y_i^{(A)}-y_i^{(B)}$ be the paired binary success difference for comparison $A$ versus $B$. We report the mean paired difference $\bar d$, a nonparametric bootstrap 95\% confidence interval, a one-tailed paired permutation test when the hypothesis is directional, and paired effect size $d_z=\bar d/\operatorname{sd}(d)$ when defined. P-values are corrected within each comparison family using Holm--Bonferroni correction.

\begin{table*}[h]
\centering
\small
\caption{Key paired statistical comparisons used in the main text. Differences are success-rate or exact-match differences, depending on the benchmark metric.}
\label{tab:key_stats}
\begin{tabular}{p{0.25\textwidth}p{0.25\textwidth}ccc}
\toprule
Benchmark & Comparison & Mean diff. & 95\% CI & Corrected $p$ \\
\midrule
Controlled drift, overall & \method{} vs. append-only & 0.167 & [0.162, 0.172] & $<0.001$ \\
Controlled drift, overall & \method{} vs. last-write-wins & 0.167 & [0.162, 0.172] & $<0.001$ \\
Controlled drift, reversal & \method{} vs. append-only & 0.740 & [0.722, 0.756] & $<0.001$ \\
Controlled drift, reversal & \method{} vs. last-write-wins & 0.740 & [0.722, 0.756] & $<0.001$ \\
Controlled drift, reversal & no memory vs. append-only & 0.098 & [0.085, 0.112] & $<0.001$ \\
Real executable, overall & \method{} vs. append-only & 0.171 & [0.165, 0.176] & $<0.001$ \\
Real executable, overall & \method{} vs. last-write-wins & 0.171 & [0.165, 0.176] & $<0.001$ \\
Real executable, reversal & \method{} vs. append-only & 0.748 & [0.728, 0.768] & $<0.001$ \\
Real executable, reversal & \method{} vs. last-write-wins & 0.748 & [0.728, 0.768] & $<0.001$ \\
Real executable, reversal & no memory vs. append-only & 0.095 & [0.083, 0.107] & $<0.001$ \\
Preference update, overall & \method{}-Full vs. append-only & 0.234 & [0.222, 0.246] & $<0.001$ \\
Preference update, overall & \method{}-Full vs. last-write-wins & 0.224 & [0.087, 0.392] & $<0.001$ \\
Preference update, reversal & \method{}-Full vs. append-only & 0.734 & [0.707, 0.760] & $<0.001$ \\
Preference update, overall & \method{}-Full vs. no memory & 0.002 & [-0.011, 0.014] & 1.000 \\
\bottomrule
\end{tabular}
\end{table*}

\paragraph{Why use paired tests.}
The main paper explains the rationale for paired tests. Here we use the same matched task or query indices for every comparison family before applying bootstrap intervals, permutation tests, and Holm--Bonferroni correction.

\section{Appendix C: Benchmark Construction and Leakage Controls}
\label{app:benchmarks}

All methods within a benchmark share the same split, query order, executor, and scoring pipeline. The evaluation suite is: controlled hidden-regime drift (50 seeds, 160 tasks per seed, phase success rate), real file-backed executable drift (50 seeds, 160 tasks per seed, phase success rate), preference-update stream (50 seeds, 160 tasks per seed, phase success rate), MemoryAgentBench SH-6k (3 query-order seeds, 300 queries, substring exact match), and MemoryAgentBench boundary settings (20--100 queries, substring exact match). This textual summary replaces a wide setup table in the main paper to avoid appendix float pages.

\begin{table}[h]
\centering
\small
\caption{MemoryAgentBench boundary settings. Values are substring exact match. Dashes indicate settings not included in the audited run.}
\label{tab:mab_boundary}
\begin{tabular}{lcc}
\toprule
Method & MH-6k & SH-262k \\
\midrule
\method{}-Rev & 0.040 & 0.000 \\
\method{}-NoRev & 0.010 & 0.000 \\
Temporal recency & 0.070 & -- \\
Append-only & 0.080 & 0.350 \\
\bottomrule
\end{tabular}
\end{table}

Table~\ref{tab:mab_boundary} reports the most space-efficient version of the boundary analysis. The omitted SH-6k and SH-32k values are already reported in the main text and MemoryAgentBench result table; the purpose here is to show the two stress conditions where the current method fails.

\begin{table}[h]
\centering
\small
\caption{Key-extraction noise audit on controlled drift. Values are full-reversal success rates over 50 seeds. Noise perturbs the structured memory key seen by memory systems, while the executable task and correct answer remain unchanged.}
\label{tab:key_noise}
\begin{tabular}{lcccc}
\toprule
Method & 0\% & 5\% & 10\% & 20\% \\
\midrule
\method{} & 0.950 & 0.908 & 0.886 & 0.777 \\
Reactive forgetting & 0.796 & 0.745 & 0.731 & 0.613 \\
Last-write-wins & 0.210 & 0.212 & 0.212 & 0.215 \\
Append-only & 0.210 & 0.212 & 0.212 & 0.215 \\
No memory & 0.309 & 0.304 & 0.302 & 0.297 \\
\bottomrule
\end{tabular}
\end{table}

Table~\ref{tab:key_noise} audits a key assumption rather than introducing a new benchmark. The task stream is unchanged, but the memory system receives a corrupted key view with probability $\rho$, quantifying how structured key errors affect the controlled reversal setting.

\paragraph{Controlled hidden-regime drift.}
The controlled benchmark is intentionally minimal. The visible context is similar across phases, but the hidden accepted tool changes. This isolates the failure mode that the paper studies: whether memory can be falsified when a previously correct precedent becomes wrong under a hidden reversal. Because the executor is deterministic, the benchmark removes language-model sampling variance and makes memory state the primary variable.

\paragraph{Real file-backed executable drift.}
The executable validation uses generated CSV/JSON files and calls real tool libraries. The accepted backend changes for CSV tasks under reversal, while JSON tasks act as stable controls. This setup guards against the criticism that the controlled benchmark is only a symbolic reward function: the agent's action is evaluated through real file I/O and code execution.

\paragraph{Preference-update stream.}
The preference benchmark separates long-term profile evidence from fresh session evidence. The stale profile remains visible to memory methods, while the current answer is determined by newer feedback.

\paragraph{MemoryAgentBench.}
MemoryAgentBench SH-6k is used as an external single-hop conflict-resolution benchmark. All compared methods run under the same query order and answer-scoring pipeline. Boundary settings, including MH-6k and very long-context variants, are reported as limits rather than as positive claims.

\paragraph{Conflict-key and value extraction.}
The main text summarizes the schema-based extractors used in each benchmark. Table~\ref{tab:key_noise} adds the key-noise audit that stress-tests this assumption in controlled drift; extending the same analysis to open-domain LLM-parsed keys is left as future work.

\paragraph{Leakage controls.}
No method receives the hidden phase label $z_t$ or the future phase boundary at decision time, except the diagnostic oracle-reset baseline. \method{} receives only the evidence stream and must infer stale evidence through conflict-keyed updates.

\section{Appendix D: Baseline Definitions and Fidelity}
\label{app:baseline_fidelity}

The paper separates two categories that are easy to conflate. \emph{Mechanism baselines} are controls implemented under the shared evaluator to test a memory principle such as append-only storage, recency, or sliding windows. \emph{Ablations} are variants of \method{} and are not counted as baselines. This separation is important because the paper's claim is about a memory lifecycle operation, not about a full deployed assistant stack.

A shared evaluator improves fairness because every method sees the same queries, answer backend, and scoring function. Table~\ref{tab:fidelity} summarizes what each mechanism baseline tests.

\begin{table*}[t]
\centering
\small
\caption{Mechanism baseline definitions. Each baseline isolates one alternative to conflict-keyed revocation under the shared evaluator.}
\label{tab:fidelity}
\begin{tabular}{p{0.18\textwidth}p{0.30\textwidth}p{0.42\textwidth}}
\toprule
Baseline & Memory operation & Purpose \\
\midrule
No memory & Retrieve no stored evidence & Measures base task difficulty without memory exposure \\
Append-only & Keep all evidence active & Tests whether stale active evidence causes pollution \\
Temporal recency & Prefer most recent memories & Tests age-based forgetting without conflict keys \\
Semantic retrieval & Rank memories by lexical relevance & Tests retrieval without lifecycle state \\
Sliding window & Keep only a fixed recent window & Tests capacity-limited forgetting \\
Reactive forgetting & Forget after detected failure & Tests global reactive reset rather than local revocation \\
Last-write-wins & Keep only the newest value for each conflict key & Strong current-key cache baseline without posterior, archive, or trials \\
Conflict-aware recency & Collapse same-key conflicts only in retrieved candidates & Tests whether retrieval-time recency is enough without persistent lifecycle state \\
Oracle reset & Reset at true phase boundary & Diagnostic upper bound with unavailable phase knowledge \\
\bottomrule
\end{tabular}
\end{table*}

\section{Appendix E: Hyperparameters and Implementation Details}

Table~\ref{tab:hyperparameters} lists the default hyperparameters and evaluation constants used unless otherwise specified in the main text.

\begin{table*}[t]
\centering
\small
\captionsetup{justification=raggedright,singlelinecheck=false}
\caption{Default hyperparameters and evaluation constants.}
\label{tab:hyperparameters}
\begin{tabular}{p{0.30\textwidth}p{0.18\textwidth}p{0.42\textwidth}}
\toprule
Quantity & Value & Role \\
\midrule
Proposal threshold $\theta_{\rm prop}$ & 3 supports & Evidence needed before proposing a precedent \\
Revocation threshold $\theta_{\rm rev}$ & 0.3 & Posterior cutoff for removing active stale precedents \\
Promotion threshold $\theta_{\rm pro}$ & 0.6 & Posterior cutoff for activating candidate precedents \\
Beta prior $(\alpha,\beta)$ & (1, 1) & Prior counts in $q(p)$ \\
\method{}-Full trial budget & 3 checks & Support, counterfactual, and contamination checks \\
Controlled/executable/preference seeds & 50 each & Random streams per drift benchmark \\
Tasks per drift seed & 160 & Four phases of 40 tasks \\
MemoryAgentBench SH-6k & 3 seeds, 300 queries & External single-hop stale-conflict validation \\
\bottomrule
\end{tabular}
\end{table*}

\paragraph{Deterministic versus model-backed settings.}
The main paper summarizes this split. In implementation, deterministic streams are used for causal memory-state tests, while MemoryAgentBench uses a fixed answer backend and prompt to test transfer to natural-language queries.

\paragraph{Scalability audit.}
Table~\ref{tab:scalability} reports the microbenchmark underlying the main-text scalability discussion. The audit uses up to one million updates and reports update cost per 1k updates plus keyed query cost.

\begin{table}[h]
\centering
\small
\caption{Scalability microbenchmark. Update cost is milliseconds per 1k updates; query cost is microseconds per key lookup.}
\label{tab:scalability}
\begin{tabular}{llrr}
\toprule
Updates & Method & Upd. & Query \\
\midrule
100k & Append scan & 0.115 & 823.971 \\
100k & Last-write-wins & 0.275 & 0.241 \\
100k & \method{} lifecycle & 0.322 & 0.216 \\
1M & Append scan & 0.114 & 4006.250 \\
1M & Last-write-wins & 0.440 & 0.399 \\
1M & \method{} lifecycle & 0.513 & 0.301 \\
\bottomrule
\end{tabular}
\end{table}

\section{Appendix F: Extended Related Work}
\label{app:extended_related}

This appendix records related areas that inform the work but are not central enough to occupy main-text space. The main text focuses on agent memory systems and stale-conflict benchmarks because they define the immediate problem. The areas below motivate design boundaries: \method{} is a memory-lifecycle mechanism, not a general retrieval architecture, a continual-learning algorithm, or a model-editing method.

\paragraph{Retrieval-augmented and long-context generation.}
Retrieval-augmented generation brings external evidence into the prompt or into model computation \citep{lewis2020retrieval,karpukhin2020dense,guu2020retrieval,borgeaud2022improving,izacard2022few,asai2024self,gao2023retrieval}. Long-context and memory-augmented Transformer work expands the amount of evidence that can be accessed directly or through non-parametric memory \citep{khandelwal2019generalization,wu2022memorizing,liu2024lost}. These methods primarily address access: whether relevant information can be found and presented to the model. Memory pollution is different because the harmful evidence can be highly relevant but obsolete. \method{} therefore changes the active memory set before retrieval, rather than proposing a new dense retriever or longer context window.

\paragraph{Concept drift and continual learning.}
Concept-drift methods detect or adapt to changes in a streaming data distribution \citep{gama2004learning,bifet2007learning,gama2014survey}. Continual-learning methods try to retain old capabilities while learning new ones, often through regularization, replay, or episodic memory \citep{kirkpatrick2017overcoming,lopez2017gradient,chaudhry2018efficient,rolnick2019experience,parisi2019continual}. These literatures motivate the idea that old evidence can become unreliable, but their update target is usually a classifier, parameter vector, or training distribution. \method{} instead updates the status of explicit textual memories that are candidates for prompting.

\paragraph{Model editing and factual updates.}
Model-editing methods update factual associations in model parameters or auxiliary editing memories \citep{meng2022locating,meng2022mass,mitchell2021fast,mitchell2022memory}. The motivation overlaps with ours: knowledge can become stale and must be revised. The layer differs. Model editing changes the model or an editor module, whereas \method{} changes the active memory state and leaves the base model fixed. This makes revocation transparent and reversible, but it also means that \method{} cannot correct parametric knowledge when no relevant evidence is retrieved.

\paragraph{Skill libraries and autonomous memory agents.}
Systems such as Voyager store executable skills for lifelong exploration \citep{wang2023voyager}, while recent surveys and frameworks describe broader memory mechanisms for autonomous agents \citep{zhang2025survey,jiang2024long,du2026memory}. These works broaden the design space of agent memory. Our experiments deliberately avoid claiming general autonomous-agent memory performance. The evidence in this paper supports a specific operation: conflict-keyed revocation for stale evidence.

\paragraph{Stale-memory validity and provenance.}
Concurrent work increasingly treats memory validity as an explicit evaluation target. STALE asks whether agents can know when stored memories are no longer valid \citep{chao2026stale}; Eywa emphasizes provenance-grounded memory and the distinction between revisable beliefs and immutable evidence \citep{joshi2026eywa}; MemoryArena studies interdependent multi-session agentic tasks \citep{he2026memoryarena}; and MemBench broadens memory evaluation dimensions for LLM-based agents \citep{tan2025membench}. These works are complementary to our study. They motivate richer evaluation of stale memories, while \method{} isolates one concrete operation--revocation of stale active evidence under a conflict key--and tests it under controlled, executable, preference-update, and MemoryAgentBench settings.

\section{Appendix G: Theoretical Notes}
\label{app:theory}

This appendix formalizes the limited theoretical claim used in the main text. The goal is not to prove that \method{} is globally optimal for all long-term memory tasks. The claim is narrower: in stale-conflict consolidation, where a newer value supersedes an older value under the same key, keeping the stale value active creates avoidable risk.

\paragraph{Key-current model.}
Consider a single conflict key $k$. At time $t$, the environment has a current value $v_t^\star$. An evidence item $x_t$ reveals a candidate value $v_t=\nu(x_t)$ for key $k$. A later item supersedes an earlier one when it has the same key and an incompatible value. The evaluation query asks for $v_t^\star$, and the loss is 0--1 error:
\[
\ell(a_t,v_t^\star)=\mathbb{1}[a_t\neq v_t^\star].
\]
Let an append-only memory state contain both a stale value $v^-$ and the current value $v^+$ under key $k$, with $v^-\neq v^+=v_t^\star$. A revocation memory state removes $v^-$ from the active set and retrieves only active current evidence for that key. We make no assumption about the internal language model except the following exposure condition: if stale evidence is retrieved together with current evidence, the downstream decision rule follows the stale evidence with probability $\eta\geq 0$.

\paragraph{Proposition 1: revocation weakly dominates under single-key supersession.}
Under the key-current model, if stale and current evidence share the same key and the current evidence is available, a retrieval policy that excludes the stale evidence has expected 0--1 loss no larger than a policy that retrieves both. If the downstream decision rule follows retrieved stale evidence with probability $\eta>0$, the dominance is strict.

\emph{Proof.}
Let $A$ denote the event that append-only retrieval exposes the stale item $v^-$ to the decision rule together with $v^+$. Conditioned on $A$, the decision rule outputs $v^-$ with probability $\eta$ and therefore incurs loss one on those cases because $v^-\neq v_t^\star$. On the complementary cases, the two policies have the same current evidence available, and the revocation policy has not removed any correct active value for key $k$. Thus the stale item contributes an additional nonnegative error term to append-only retrieval and contributes no additional correct information under the key-current assumption. The expected loss of append-only retrieval is therefore at least the expected loss of retrieval after revocation, with strict inequality whenever $A$ occurs with positive probability and $\eta>0$. \hfill$\square$

\paragraph{Proposition 2: pollution scales with stale exposure.}
Let $q=\Pr(A)$ be the probability that append-only retrieval exposes stale evidence for key $k$, and let $\eta=\Pr(a_t=v^- \mid A)$ be the probability that the downstream decision follows the stale evidence once exposed. If revocation leaves the current value available, then the excess risk of append-only memory over revocable memory satisfies
\[
\Risk_{\rm append}-\Risk_{\rm revoke}\geq q\eta .
\]
More generally, if following stale evidence increases expected task loss by $\Delta>0$, then
\[
\Risk_{\rm append}-\Risk_{\rm revoke}\geq q\eta\Delta .
\]

\emph{Proof.}
The event $A$ occurs with probability $q$. Conditional on $A$, stale following occurs with probability $\eta$. Each stale-following event adds loss $\Delta$ relative to a decision that uses the current value. Taking expectation gives the lower bound. The 0--1 case is recovered by setting $\Delta=1$. \hfill$\square$

This proposition supplies the formal excess-risk statement underlying the main-text intuition about stale exposure.

\paragraph{Proposition 3: posterior revocation is consistent within a stationary segment.}
Let a precedent $p$ receive Bernoulli feedback $Z_i\in\{0,1\}$ within a stationary segment, where $Z_i=1$ means the precedent is supported and $\mathbb{E}[Z_i]=\mu$. With Beta prior $(\alpha,\beta)$, the posterior mean after $n$ feedback events is
\[
q_n(p)=\frac{\alpha+\sum_{i=1}^{n} Z_i}{\alpha+\beta+n}.
\]
For any revocation threshold $\theta_{\rm rev}$, if $\mu<\theta_{\rm rev}$ then $q_n(p)<\theta_{\rm rev}$ eventually with probability one; if $\mu>\theta_{\rm rev}$, then the probability of false revocation after sufficiently many observations decays exponentially in $n$.

\emph{Proof sketch.}
By the strong law of large numbers, $n^{-1}\sum_{i=1}^{n}Z_i\rightarrow\mu$ almost surely. Since the prior contribution $(\alpha+\beta)/n$ vanishes, $q_n(p)\rightarrow\mu$ almost surely. Therefore, for $\mu<\theta_{\rm rev}$, the posterior mean eventually falls below the threshold almost surely. For $\mu>\theta_{\rm rev}$, choose $\epsilon=(\mu-\theta_{\rm rev})/2$. A false revocation event after the prior becomes negligible implies that the empirical mean deviates below $\mu-\epsilon$. Hoeffding's inequality bounds this probability by $\exp(-2n\epsilon^2)$ up to constants from the finite prior. \hfill$\square$

\paragraph{Scope of the theory.}
The propositions justify only stale single-key conflict revocation. They do not claim guarantees for retrieval-chain construction, compositional answering, or long-context selection.

\section{Appendix H: Reproducibility Notes}

Appendix~\ref{app:benchmarks} defines benchmark construction and leakage controls, Appendix~\ref{app:baseline_fidelity} defines method controls, and Table~\ref{tab:hyperparameters} collects run constants. The statistical protocol is given in Appendix B.

\clearpage
\end{document}